\documentclass{sig-alternate-05-2015}
\usepackage{xcolor} 
\makeatletter
\def\@copyrightspace{\relax}
\makeatother

\begin{document}

\setcopyright{acmcopyright}

\doi{10.475/123_4}

\isbn{123-4567-24-567/08/06}

\conferenceinfo{PLDI '13}{June 16--19, 2013, Seattle, WA, USA}

\acmPrice{\$15.00}

%
\conferenceinfo{WOODSTOCK}{'97 El Paso, Texas USA}

\title{Realizing Uncertainty-Aware Timing Stack in Embedded Operating System
}
%
%
%
%
%

\numberofauthors{3} 
%
\author{
%
%
\alignauthor
\large Amr Alanwar, Fatima M. Anwar\\
       \affaddr{University of California,}\\
      \affaddr{Los Angeles}\\
\alignauthor
\large Jo{$\tilde{a}$}o P. Hespanha\\
       \affaddr{University of California,}\\
       \affaddr{Santa Barbara}\\
\alignauthor
\large Mani B. Srivastava\\
       \affaddr{University of California,}\\
       \affaddr{Los Angeles}\\
\additionalauthors{dsdsd,
email: {\texttt{jsmith@affiliation.org}}) and Julius P.~Kumquat
(The Kumquat Consortium, email: {\texttt{jpkumquat@consortium.net}}).}
\date{30 July 1999}
}
\additionalauthors{Additional authors: John Smith (The Th{\o}rv{\"a}ld Group,
email: {\texttt{jsmith@affiliation.org}}) and Julius P.~Kumquat
(The Kumquat Consortium, email: {\texttt{jpkumquat@consortium.net}}).}
\date{30 July 1999}

\maketitle\begin{abstract}


Time synchronization has been studied extensively over the recent years with an advent of time critical applications for wireless sensor networks. The distribution of the global reference time over the radio links is the most popular synchronization mechanism. Often overlooked, the timing uncertainties spread across transmitter to the receiver, limit the accuracy of state-of-the-art synchronization protocols. These timing uncertainties are due to the instability of crystal oscillators and timestamping mechanisms. The effect of these uncertainties is cumulative in nature and build up the synchronization error. On the other hand, limited resources of energy, computational units, storage, and bandwidth are a driving force towards lightweight protocols. Hence, this paper presents a deep analysis of each source of timing uncertainty and motivates uncertainty-aware time synchronization. Extensive experiments are conducted to highlight the contribution of each source and provide recommendations for mitigating the resultant timing uncertainty. We have also proposed a Lightweight Kalman filter based on our analysis to meet the limited resources constraint.
\end{abstract}

\section{Introduction}

\let\thefootnote\relax\footnote{This research is funded in part by the National Science Foundation under awards  CNS-1329755 and CNS-1329644. The U.S. Government is authorized to reproduce and distribute reprints for Governmental purposes notwithstanding any copyright notation thereon. The views and conclusions contained herein are those of the authors and should not be interpreted as necessarily representing the official policies or endorsements,
either expressed or implied, of NSF, or the U.S. Government.}
\let\thefootnote\relax\footnote{Authors emails are $\{$alanwar,fatimanwar,mbs$\}$@ucla.edu, and hespanha@ece.ucsb.edu.}

\let\thefootnote\relax\footnote{EWiLi\rq 16, October 6th, 2016, Pittsburgh, USA. Copyright retained by the authors.}

\vspace{-2em}

Awareness of time is critical to the reliability, security, and performance of distributed systems. 
Furthermore, traditional clock models do not cover all sources of uncertainties and synchronization protocols provide fixed accuracy with no visibility into the resultant uncertainty, even though it has proven that the knowledge of uncertainty in time enhances performance and reduces system complexity. For example, Google Spanner \cite{spanner} uses uncertainty in time to achieve external consistency of global database transactions. For real-time interactivity in cloud gaming, Outatime \cite{outatime} relaxes the stringent demands on frame speculation, provided the uncertainty in time is less than 100 milliseconds.

Time synchronization is the process of bringing an entire network to one common reference time. This reference could be the local clock time of a node in the network or a third party device with an accurate sense of time. In most cases, a reference node (often called the root node) in the network is considered to have the global time. The root and all the nodes in the network timestamp a common event with their local clocks. The root broadcasts the global timestamp of that common event in the network. Every client node receives the global timestamp and pairs it with local timestamp of that event. Sequence of such common events provides global and local timestamp pairs. Finally the clock adjustment mechanism depends on these pairs to provide a mapping between the local and global time. This mapping is used to adjust the local clock at every client node. However, as simple as this process may sound, it suffers from many sources of uncertainties that limit the capabilities of synchronization protocols. State-of-the-art synchronization protocol~\cite{pulsesync} have taken clock drift as the only source of uncertainty in timing and reported the resultant uncertainty between the transmitter and receiver. In contrast, we have pointed out some other sources of uncertainty that have an equal, if not more, effect on the synchronization accuracy. We conducted detailed experiments to show the contribution of each and every source of timing uncertainty from the transmitter to the receiver. Therefore, we argue that better synchronization can be achieved by considering all the sources of uncertainties and mitigating their effects.

\begin{figure}[th]
\centering
\includegraphics[scale=0.45, angle=0]{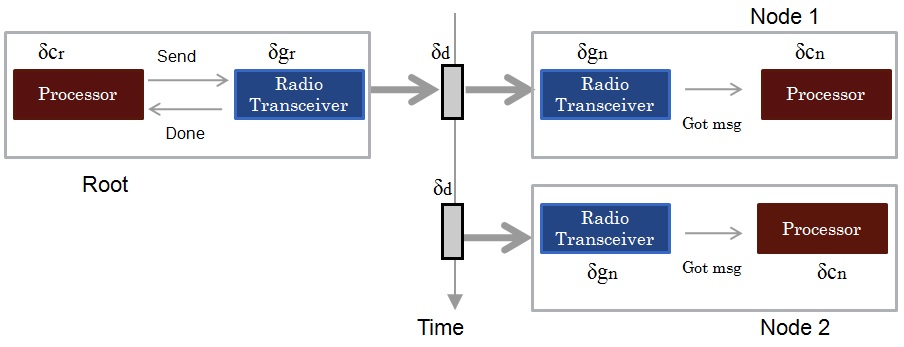}
\caption{Uncertainty in a synchronization process}
\label{fig:uncer}
\end{figure}

Hardware capabilities have evolved over recent years to remove certain sources of uncertainties. For example, MAC-layer time stamping on wireless network greatly reduced timing uncertainty. Also, many radio transceivers now support interrupts at transmission and reception times. Certain microcontrollers have timer capture capability to timestamp radio transceiver interrupts. Despite these hardware improvements, most timestamping mechanisms still suffer from timing jitter due to various uncertainties involved, preventing better synchronization accuracy. Main sources of uncertainty are presented in Fig.~\ref{fig:uncer} and are summarized through the following points:

\begin{itemize}
  \item Clock drift: instability of the crystal oscillators causes relative clock drift between the root and the network nodes. All the node clocks are continuously drifting in time with respect to the root clock. 
  \item Radio transceiver interrupt generation: the capability of radio transceivers to generate an interrupt on the precise time can be a bottleneck in achieving a common notion of time. We denote $\delta_{g_{r}}$ and $\delta_{g_{n}}$ as radio transceiver interrupt generation uncertainty at the root and the node, respectively.
  \item Interrupt capture: wireless nodes use input capture hardware module to record the timestamp of an input signal. The crystal type and frequency, which drives the timer, affect the accuracy of the timestamp value. We denote $\delta_{c_{r}}$ and $\delta_{c_{n}}$ as interrupt capture uncertainty at the root and the node, respectively.  
  \item Propagation delay: communication channel types and the distance between the root and the node affect the propagation delay uncertainty $\delta_{d}$.
\end{itemize}

We analyze the effects of the above sources of uncertainty on overall timestamping capability. Through our analysis, we extract a set of recommendations to enhance the overall timing accuracy. In addition, we propose a LightWeight Kalman filter (LW Kalman) for time synchronization on a single hop network topology based on these recommendations. LW Kalman compensates for the uncertainties described above and achieves higher synchronization accuracy without sacrificing the limited resources available on most wireless devices. In the next section, we describe and analyze our new clock model, and in section \ref{sect:tune-time-sync}, we explain the LW Kalman synchronization protocol.

\section{Clock Model}
\label{sect:model}

Let $k$ be the common event time index (synchronization message time index). We denote the average relative clock drift between the master (root) and the slave (node) as $\phi$ and the uncertainty around $\phi$ as $\tilde{n}_r(k)$, where $\tilde{n}_r(k)$ has zero mean. Also, $\bigtriangleup_{nr}$ is relative constant offset between the root and the node at k = 0.
\\Time synchronization mainly depends on generating and capturing radio transceiver interrupts at the common event. 
The network nodes get a noisy timestamps of the common event due to uncertainties. We denote ${R}$ and ${N}$ as the noisy timestamps of the root and the node, respectively. We assume that the root timestamps the event time perfectly and all the noise is at the node side. The validity of the previous assumption will be shown later. Both the root and node have the notion of the common event time as shown in equation~\ref{equ:capt}, where $\bigtriangleup_{d}$ is the average propagation delay between the root and node, and $\delta_d(k)$ is zero mean uncertainty associated with the propagation delay. $\delta_{g_{nr}}(k)$ is the relative uncertainty in generating interrupt at the radio transceivers, and $\delta_{c_{nr}}(k)$ is the relative uncertainty in capturing the timestamp of the interrupt on both sides.

\begin{equation}
\label{equ:capt}
\begin{split}
R(k)  &= k \\
N(k)  &= k + \phi*k + \tilde{n}_r(k) + \bigtriangleup_{nr}  + \bigtriangleup_{d}  + \delta_d(k) \\
	  &+ \delta_{g_{nr}}(k) + \delta_{c_{nr}}(k) 
\end{split}
\end{equation}

Message based time synchronization protocols depend on sending synchronization messages every predefined period, 
and analyzing a sequence of root and node timestamps can give a better understanding of the introduced uncertainty. We denote $s(k)$ as the instantaneous rate of change in timestamps between root and node, as shown in equation~\ref{equ:slope}.

\begin{equation}
\label{equ:slope}
\begin{split}
s(k) &= \frac{N(k) - N(k-1)}{R(k) - R(k-1)} \\
\end{split}
\end{equation}

Substituting equation~\ref{equ:capt} in equation~\ref{equ:slope} results in the following:
\begin{equation}
\label{equ:slope_sub}
\begin{split}
s(k) &= k + \phi*k + \bigtriangleup_{nr} + \tilde{n}_r(k) + \bigtriangleup_{d}  + \delta_d(k) \\
     &+\delta_{c_{nr}}(k) + \delta_{g_{nr}}(k) \\
     &-[\ k -1 + \phi*(k - 1) + \bigtriangleup_{nr} + \tilde{n}_r(k-1)  \\
     &+ \bigtriangleup_{d}  + \delta_d(k-1)  + \delta_{c_{nr}}(k-1) + \delta_{g_{nr}}(k-1)\ ]\\
     \\
     &= 1 + \phi + \bigtriangleup{\tilde{n}_r(k)} + \bigtriangleup{\delta_{d}}(k) + \bigtriangleup{\delta_{c_{nr}}(k)} + \bigtriangleup{\delta_{g_{nr}}(k)}
\end{split}
\end{equation}

\begin{figure}[tbp]
\centering
\includegraphics[scale=0.3, angle=0]{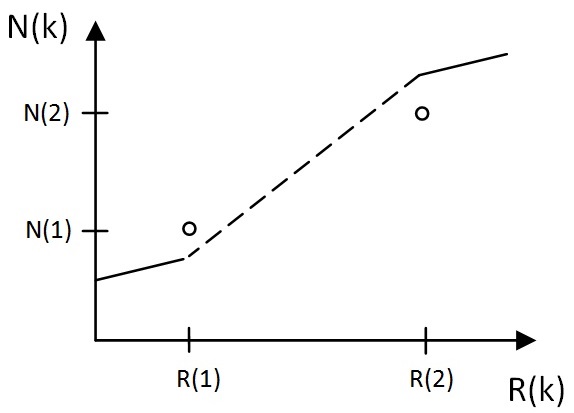}
\caption{Relation between root and node time}
\label{fig:lin_mod}
\end{figure}

 
Generally, a linear model is assumed between root and node in each synchronization period as shown in Figure~\ref{fig:lin_mod}. The mean, standard deviation and variance of $s(k)$ are $m_s$, $std_s$, and $var_s$, respectively, where accuracy of this model depends mainly on $var_s$ of the rate of change.
Higher noise in timestamps results in higher values of uncertainties in the clock model. 

We now quantify different kind of uncertainties mentioned in our model through experimental analysis.

\subsection{Experimental Analysis}
We use Beaglebone Black (BBB) -- an embedded platform loaded with an operating system -- and interface it with AT86RF233, which is a low power, 2.4GHz Zigbee Transceiver packaged under Atmel AT86RF233 Evaluation Kit (ATREB233SMAD-EK).
We configure one Transceiver in transmit mode (TX) and the other in receive mode (RX). Note that the Atmel transceivers provide interrupts on a GPIO pin at the start and end of frame transmission and reception.
BBB is interfaced to the transceiver, captures its internal timer upon TX or RX interrupt. The BBB timer's clock source is chosen to be pre-scaled versions of transceiver's board clock (2Mhz, 4Mhz and 16Mhz), whereas the transceiver itself is always clocked at 16MHz. We compare $m_s$ and $var_s$ of $s(k)$ in equation~\ref{equ:slope_sub} for different kinds of uncertainties. 

 
\subsubsection{Uncertainty in generating transceiver interrupt}
Variety of radio transceivers, nowadays, offer interrupts at the start and the end of frames. However, these interrupts are not deterministic and have an uncertainty in their generation, $\delta_{g_{nr}}(k)$. We have divided our experiment into two parts.

\begin{figure}[tbp]
\centering
\includegraphics[width=1\columnwidth]{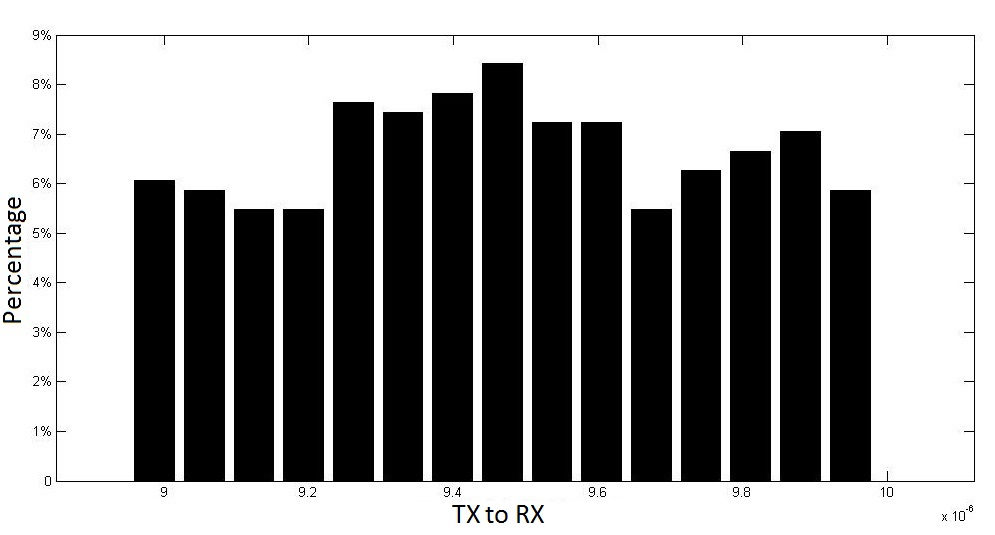}
\caption{Histogram of TX to RX interrupt generation uncertainty, using WaveRunner oscilloscope}
\label{fig:Tx_Rx_fal}
\end{figure}

\textit{TX to RX interrupt generation uncertainty:}
The BBB records timestamps for the falling edges of the interrupts, both at TX and RX. We calculate $\delta_{g_{nr}}(k)$ by subtracting the maximum difference of TX and RX timestamps from the minimum difference of TX and RX timestamps. This uncertainty has a uniform distribution as shown in Figure \ref{fig:Tx_Rx_fal} and is uncorrelated to time. Our experiments show an uncertainty of 1.031$\mu$sec and a $var_s$ of 1.892e-15 in $s(k)$ estimation. $\delta_{g_{nr}}(k)$ contributes by 1.031$\mu$s to the offset estimation, which is about two timer ticks for 2Mhz clock. Generally speaking, its contribution is ( $1.031\mu s * clock freq$ ) timer ticks error for each timestamp, and its effect is cumulative in nature.

\begin{figure}[htbp]
\centering
\includegraphics[width=1\columnwidth]{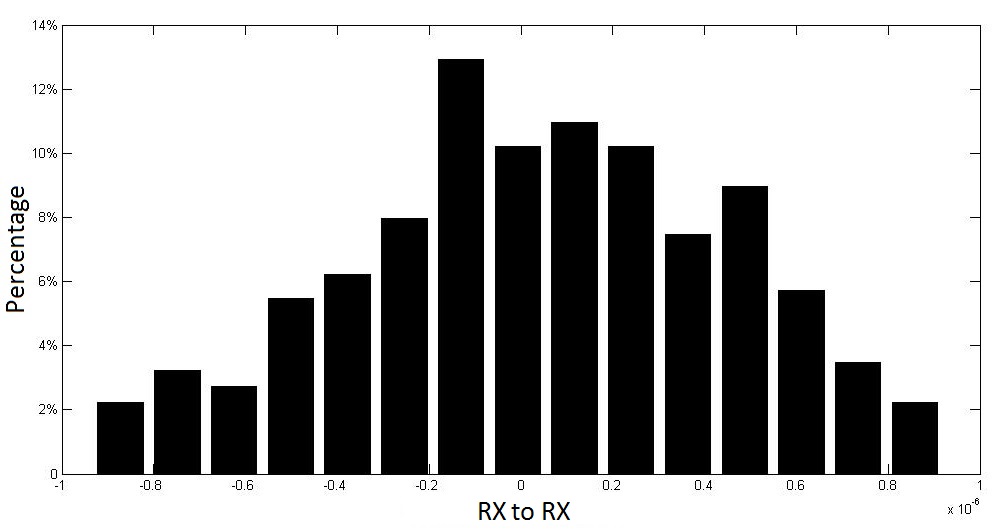}
\caption{Histogram of RX to RX interrupt generation uncertainty, using WaveRunner oscilloscope}
\label{fig:Rx_Rx}
\end{figure}

\textit{RX to RX interrupt generation uncertainty:}
We configure three radio transceivers (one TX and two RXs) to measure RX to RX uncertainty. $\delta_{g_{nr}}(k)$ is 1.854$\mu$s with a mean value of 0.037408$\mu$s as shown in Figure \ref{fig:Rx_Rx}, which is almost double the uncertainty between TX to RX. Therefore, our results suggest that the reference broadcast \cite{rbs} is not helpful to decrease $\delta_{g_{nr}}(k)$. However, this behavior may change in wide wireless networks where the distance between the nodes is longer and the uncertainty in the propagation delay contributes more as compared to $\delta_{g_{nr}}(k)$. Comparing the RX to RX distribution with the TX to RX gives us an understanding that most of the noise is at the RX side as the uncertainty has almost doubled and the two RX resultant distributions are approximated to a triangular distribution shown in Figure \ref{fig:Rx_Rx}. These results support our initial assumption of considering all the noise at the slave node (RX). This gives an intuition to avoid reference broadcast based RX to RX synchronization for better accuracy.

\begin{figure}[bp]
\centering
\includegraphics[width=1\columnwidth]{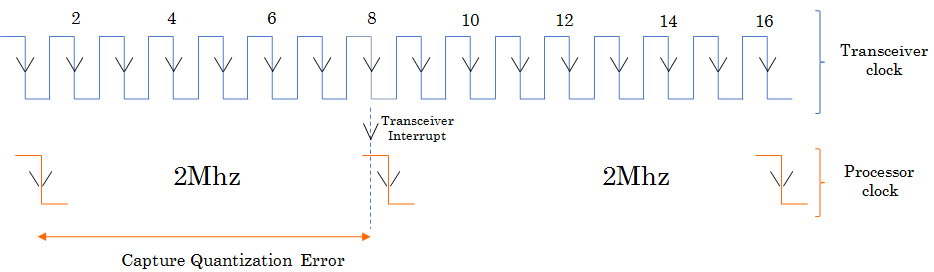}
\caption{Quantization error in input capture}
\label{fig:capt}
\end{figure}


\subsubsection{Uncertainty in capturing interrupt}
Processors use timer capture to get accurate timestamps upon a TX and RX interrupt. This capture is done with a hardware support to get rid of the software timer reading uncertainty. We define two types of interrupt capture, first one is the \textit{asynchronous capture}, where the processor timer and the radio transceiver are clocked using different crystals. The second type is the \textit{synchronous capture}, where the processor timer and the transceiver are clocked by the same crystal.

In commodity platforms, the transceiver's interrupt is always generated in synchronous with the transceiver's clock, while the processor timer captures this interrupt using another clock/frequency. Therefore, a capture quantization error is introduced as shown in Fig.~\ref{fig:capt}. The effect of capture quantization error is one timer tick for each captured value in the worst case. However, this effect grows when processing multiple timestamps over multiple synchronization periods. The quantization error and the interrupt capture type contribute to the relative interrupt capture uncertainty $\delta_{c_{nr}}$. 

\begin{table}[t]
\caption{Effect of increasing the frequency on capture quantization error}
\centering
\normalsize
\begin{tabular}{l||ccc}
 Frequency &  $m_s$	&	$var_s$ \\
\hline\hline
      ~2Mhz	&	1 + 1.2690e-07				&	3.2733e-15 \\

   	  ~4Mhz	&	1 + 1.0447e-07				&	2.2599e-15  \\

  	  ~16Mhz  &	1 + 3.2476e-08				&	1.4543e-15 \\
  		 	 
\end{tabular}
\label{tab:capt}
\end{table}

The capture quantization error can be removed by timestamping the radio interrupts at the same frequency as they are generated. Table~\ref{tab:capt} shows the decreased values of $var_s$ and $m_s$ at 16MHz as compared to other frequencies. This is because the radio is generating interrupts in synchronous to 16MHz clock, and the processor is capturing these interrupts synchronously with the same 16MHz frequency. This is an example of synchronous capture. Unfortunately, this is not the case in many boards, where the processor and radios are clocked by different clock frequencies.
One way to counter the quantization error is to make the processor timestamp the interrupts twice using its rising and falling edge. This double sampling over short period enhances the results.



Now we show that there is a correlation between $\delta_{c_{nr}}$ and the clock used in generating the interrupt. We compare synchronous versus asynchronous capture where all crystals are clocked at 16MHz.
We conduct three experiments to analyze this uncertainty. First experiment ($exp1$) looks into synchronous capture where each BBB is clocked by its own transceiver clock. Then, we swap the clock sources in second experiment ($exp2$) i.e, the TX BBB is clocked by the RX transceiver clock and vice versa. The goal of this experiment is to have asynchronous capture without introducing any new clock drift to the equation~\ref{equ:slope_sub}. Finally, we use external clock sources to clock both TX and RX BBB to have asynchronous capture in the third experiment ($exp3$). The external clock sources are Siward (SX-4025), which is of the same type as the transceiver clocks. Therefore, we have a fair comparison across all the experiments.

We conduct and compare these experiments at high frequency (16Mhz) to remove capture quantization uncertainty. Interestingly, introducing external clock sources in the asynchronous capture ($exp3$) has the worst performance while the synchronous case ($exp1$) produces the best results as shown in Table \ref{tab:cap_16Mhz}. $exp2$ shows that the asynchronous capturing without external clocks is better than $exp3$, but worse than $exp1$. Therefore, we conclude that the correlation between the event capturing clock frequency and the event generating clock frequency is helpful in decreasing the overall variance. On the other hand, introducing new clock sources increases the overall $var_s$.

\begin{table}[tbp]
\caption{Analysis of 16Mhz input capture type}
\centering
\normalsize
\begin{tabular}{l||cc}
  &  $m_s$	&	$var_s$\\
\hline\hline
     $exp1$ &	1 + 3.2476e-08				& 1.4543e-15	 \\
      $exp2$ & 	1 + 5.4306e-07 				& 2.5578e-15	\\
  $exp3$ & 1+  7.5548e-06  	& 3.3785e-15\\		 

\end{tabular}
\label{tab:cap_16Mhz}
\end{table}

Also, it is worth mentioning that the transceiver crystals are much more stable than processor crystals on most embedded platforms. On the other hand, the cost of keeping the transceiver awake all the time just for the sake of supplying the clock to the processor timer is quite high. Therefore, we propose a tradeoff where a processor choose transceiver clock on the transceiver awake time and switch to its own clock on the transceiver sleep time. However, this configuration needs to internally synchronize different clock sources before switching. The summary of this section is that synchronous capture is better than the asynchronous one. Also, synchronous capture on the highest frequency is the best for timestamping accuracy.

\subsubsection{Relative clock drift}
\begin{figure}[tbp]
\centering
\includegraphics[scale=0.4, angle=0]{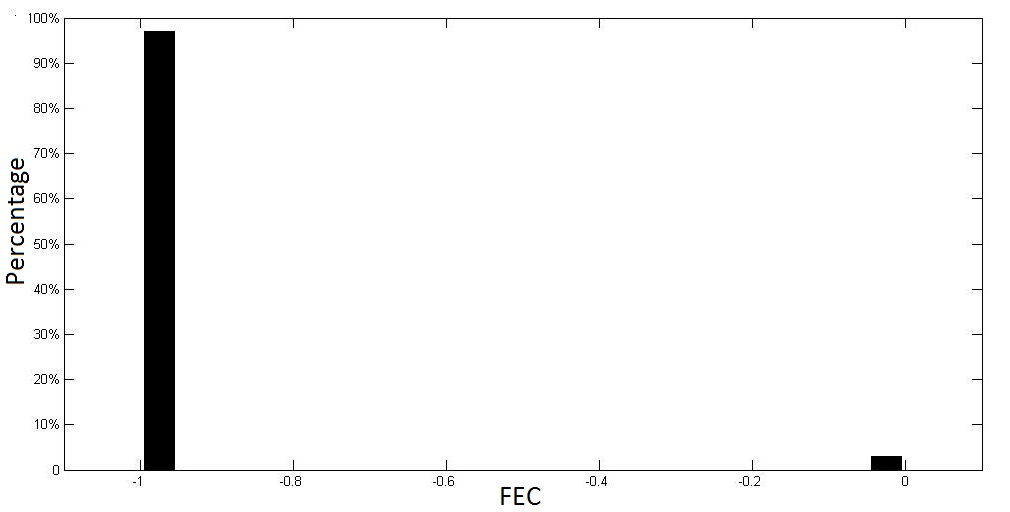}
\caption{FEC histogram over 500 samples}
\label{fig:FEC}
\end{figure}

Comparing the frequency of a signal generated by the root and node clocks can be used to determine the relative clock/crystal drift. In some transceivers, Instantaneous Frequency Error Calculation (FEC) between the root and node is calculated upon every received frame to measure the relative drift. 
We collect 500 samples of FEC at RX and calculate the relative crystal drift ($f_{crystal}$) between TX crystal and RX crystal using the equation, ($f_{crystal}[ppm] = FEC * ( 5e5 /128) / ( f_{RF}\textrm{[MHz]})$), where $f_{RF}$ is 2405Mhz in our experiment. FEC is an 8-bit register represented as a two's complement signed value. Figure~\ref{fig:FEC} shows the histogram of our 500 samples of FEC. Almost 97\% the FEC sample values are -1, and only 3\% of the values give zero relative offset. The average FEC value is 0.9702. This results in $f_{crystal}[ppm] = 1.5702$. We can relate the crystal drift $f_{crystal}$ to $m_s$ of $s(k)$ using, $m_s = 1 + f_{crystal}[ppm]/1e6$.

FEC results give a proof that the crystal drift affects $m_s$ and does not have high contribution to $var_s$ as 97\% of the FEC samples give the same value. Better relative drift can be calculated by introducing frequency hopping, which gives more accurate FEC values.
Therefore, we suggest to improve the accuracy of this feature in the transceiver to be used in the synchronization models.

\subsubsection{MAC software package jitter}
The timestamp provided by many platforms in wireless networks is based on n-bits hardware counter concatenated with n-bits software variable. PulseSync with tinyos~\cite{pulsesync} and Atmel MAC Software package use 16-bit hardware counter combined with 16-bit software variable. Overflows should be handled carefully to avoid missing interrupts. This adds several limitation on the whole system. For example, if a processor works on 16Mhz, an overflow interrupt will fire almost every 4.096ms which increases the risk of using critical sections and introduces high amount of overhead, thus affecting $m_s$. Also, there is a probability of having a race condition when an interrupt shows up at the overflow edge.

\subsubsection{Operating System Uncertainty}
The overhead of the Linux kernel contributes significantly to the end-to-end uncertainty associated with reading a timestamp. This uncertainty varies, and is a function of different factors, such as the system load and CPU operating frequency. A probabilistic estimate of this uncertainty in a given time window can be made by reading the timestamp in a tight loop from userspace. By taking the difference of consecutive timestamps in a given time window, we can calculate the distribution of the OS uncertainty associated with reading the clock. 

\section{QoT-Aware Time Synchronization}
\label{sect:tune-time-sync}

After describing various sources of uncertainties, we implement a synchronization protocol that is Quality of time (QoT) aware; not only is it controllable but also provides an observable uncertainty, thus giving an insight into the achieved performance to the application. 
The time synchronization service is aware of the underlying clock model and is able to control its parameters to achieve the required synchronization accuracy.

We propose a Kalman filter based time synchronization protocol that makes use of the uncertainties exposed by the clock model. Kalman filter is one of the most popular techniques to estimate unknown parameters. It addresses the problem of state estimation from actual measurements. It is a statistically good estimator for noisy measurements with a gaussian distribution. We implement a Lightweight version of scalar Kalman for time synchronization to meet sensor networks limited resource constraints. We assume a synchronization approach which uses one-way packet exchange. This way the average propagation delay can be estimated beforehand~\cite{ftsp}.

\subsection{Light Weight Kalman Filter}
The frequency offset $f_o(k)$ defines the relative clock frequency drift between the root and the node within the synchronization period and it is assumed to be constant within one period. $f_o(k)$  is calculated using equation~\ref{equ:foff}. We denote the mean and uncertainty in timestamp values as $\mu_{q}$ and $e_{q}$ respectively. $e_{q}$ accounts for the uncertainties in the transceiver interrupt generation, interrupt capture and the propagation delay as described in section \ref{sect:model}. The next state of $f_o(k)$ is modeled by equation~\ref{equ:foff2}, where $\mu_{rd}$ and $e_{rd}$ are the mean and uncertainty in relative clock drift respectively. LW Kalman has the capability of compensating for different sources of uncertainty.

\begin{equation}
\label{equ:foff}
\begin{split}
f_o(k) &= \frac{\bigtriangleup{R(k)} - \bigtriangleup{N(k)}}{ \bigtriangleup{N(k)}} + \mu_q + e_q
\end{split}
\end{equation}

\begin{equation}
\label{equ:foff2}
\begin{split}
f_o(k+1)&= f_o(k) + \mu_{rd}+ e_{rd}
\end{split}
\end{equation}

where 
\begin{equation}
\begin{split}
\bigtriangleup{R(k)}  &= R(k) - R(k-1)\\
\bigtriangleup{N(k)}  &= N(k) - N(k-1)
\end{split}
\end{equation}

 
 Each node has an estimated global (root) time $N_g$, which is the output of the synchronization model at the node. The synchronization error at the node at any time $m$ is $sync\_err(m) = R(m) - N_g(m)$, where $R(m)$ is the actual global (root) time at $m$. If we have an estimated frequency offset $\hat{f_o}(k)$, then $N_g(m)$ can be calculated within any time synchronization period using equation~\ref{equ:ng}.

\begin{equation}
\begin{split}
\hat{f_o}(k)&= \frac{\bigtriangleup{R(k)} - \bigtriangleup{N(k)}}{ \bigtriangleup{N(k)}} \\
\bigtriangleup{R(k)} &= \bigtriangleup{N(k)} + \hat{f_o}(k) * \bigtriangleup{N(k)}   
\end{split}
\end{equation}

Then at any time index m, where $k < m < k+1$,
\begin{equation}
\label{equ:ng}
N_g(m)= R(k) + ( N(m) - N(k)  )* ( \hat{f_o}(k) + 1 ) 
\end{equation}

Previous work \cite{conf:kalmit} assumed the measurement error to be zero and applied Kalman filter to the time stamped messages between master and slave computers. Their results were compared to NTP protocol. Another work~\cite{conf:kalchin} implemented multi-dimensional Kalman filter at 8Mhz for 36 minutes. LW Kalman reports better results at only 2Mhz for 1 hour after choosing the uncertainty covariance matrix properly.


We choose $f_o(k)$ to be the state x of Kalman filter. $f_o(k)$ equals s(k)-1 to avoid expensive floating point calculation. The state transition scalar $A$ and measurement scalar $H$ equals one. The input control scalar $B$ is equal to zero. We calculate corrected state $\hat{x}'(k)$ by finding a weighted difference between the actual measurement $z(k)$ and the predicted measurement. We define $Q$ and $\tilde{R}$ as the model and the measurement noise variance, respectively. The time and measurement update are as following : 
\begin{enumerate}
\item{Project the state and error covariance ahead}

\begin{equation}
\label{equ:kal}
\begin{split}
\hat{x}'(k) &= \hat{x}(k-1) \\
\hat{P}'(k) &= \hat{P}(k-1) + Q
\end{split}
\end{equation}

\item{Compute the Kalman gain}
\begin{equation}
K(k) = \frac{\hat{P}'(k)}{\hat{P}'(k) + \tilde{R}} 
\end{equation}
\item{Update state estimation and error covariance based on new measurement z(k)}
\begin{equation}
\begin{split}
\hat{x}(k) &= \hat{x}'(k) + K(k)*( z(k) - \hat{x}'(k ) \\
\hat{P}(k) &= ( 1 - K(k) ) * \hat{P}'(k)
\end{split}
\end{equation}
\end{enumerate}

\subsubsection{Evaluation}
We implement LW Kalman and FTSP \cite{ftsp} on the application layer of the MAC layer software package in Atmel AT86RF233 Evaluation Kit (ATREB233SMAD-EK). We should note that authors in \cite{pulsesync} depend on FTSP for single hop synchronization.
We make sure that LW Kalman and FTSP mathematical models are updated in parallel for a fair comparison. We compare their synchronization errors over different synchronization period. We make use of two message types in our system which are as follows:

\begin{enumerate}
\item Synchronization message : This is a normal message in which the root sends its time to node so that it can update its mathematical models (both FTSP and LW Kalman models) using the new reading. Its period is the synchronization period.

\item Query message : This message is used to test the synchronization between the root and the node. We configured the root to send this message every 18 seconds. FTSP~\cite{ftsp} choose a third party node to do this job and act as a reference broadcaster between the root and node. However, the uncertainty between the TX and RX is less than the uncertainty between RX and RX (established in section \ref{sect:model}). Therefore, we configure the root to send the query message in order to have the best confidence level in our testing methodology as done in \cite{pulsesync}.
\end{enumerate}

\begin{table}[t]
  \centering
  \caption{Synchronization error mean and standard deviation for LW Kalman and FTSP}
\centering
\normalsize
    \begin{tabular}{l||c l||l c}
    & \multicolumn{2}{c}{FTSP}  & \multicolumn{2}{c}{LW Kalman } \\
  Sync. Period			  & mean  & std   & mean  & std   \\
\hline\hline
\centering
    30sec    & + 0.349   & 0.611   & + 0.160& 0.587  \\
    1min     & - 0.139 & 1.097  & - 0.104& 0.769 \\
    3min     & - 0.817 & 2.782    & - 0.388 & 1.432     \\
    6min     & + 0.965  & 4.623     & + 0.495 & 3.731     \\
    \end{tabular}
  \label{tab:kal_ftsp_m_Std}
\end{table}

\begin{figure}[tbp]
\centering
\includegraphics[height=5cm]{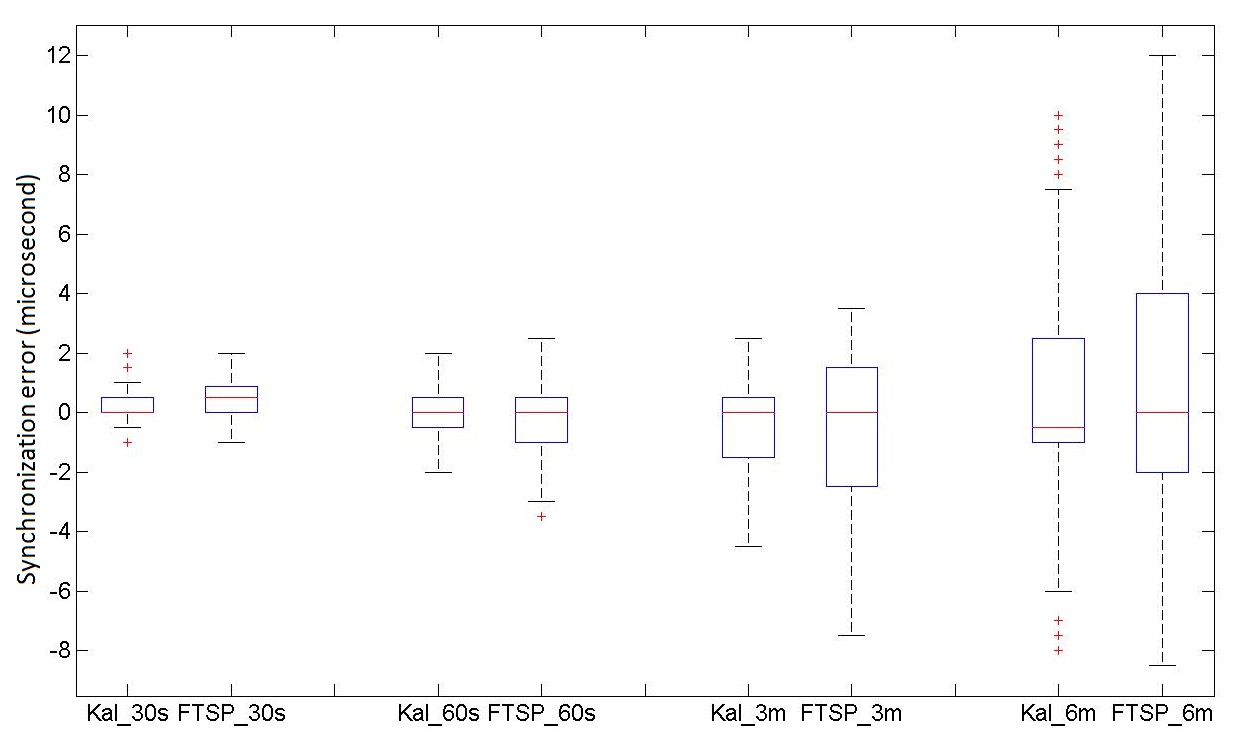}
\caption{LW Kalman and FTSP boxplot for different synchronization periods}
\label{fig:boxplot}
\end{figure}

LW Kalman has a better mean and standard deviation as compared to FTSP as shown in Table~\ref{tab:kal_ftsp_m_Std}. We know that short synchronization period costs more bandwidth and power as it prevents the transceiver to go into deep sleep mode. A synchronization protocol should be able to tune its performance by choosing a longer synchronization period for an application that does not require a very high synchronization accuracy, thus conserving resources. We investigate the stability of LW Kalman and FTSP over longer synchronization periods, and we find that 
LW Kalman is more stable over longer periods. Also, Fig.~\ref{fig:boxplot} visualizes the comparison between the two protocols over different synchronization periods. The disadvantage of LW Kalman is that it needs offline training to calculate the best convaraince values depending upon the synchronization period. 
However, this offline training is only done once for each synchronization period.

LW Kalman source code on top of the modified Atmel MAC software package and the image used in BBB can be found in~\cite{conf:oursrc}.
\section{Related Work}
\label{sect:related-work}
The notion of time uncertainty is not new as NTP \cite{ntp} computes an uncertain bound on time; it does not expose this bound to an application however and it becomes invalid when a clock adjustment is made.

In recent literature, many time synchronization techniques use analytical modeling. 
Adaptive Clock Estimation and Synchronization (ACES)~\cite{aces} models the clock directly and applies Kalman filter to track the clock offset and skew. 
In~\cite{conf:kalmit}, clock is synchronized over the packet switched network using Kalman filtering. But their assumption of constant clock skew over a long time is unrealistic for off-the-shelf unstable clocks. 
Seong et al.~\cite{18} identified quantization error in timestamping and compensated for this error using feed forward filter preceding a PI controller. Xu et al.~\cite{13} uses a Kalman filter based proportional-integral (PI) clock servo to correct for this quantization error and clock offset in cascaded real time sensor networks. However, no comparison against an existing optimal PI controller has been given.
\section{Conclusion}
With the advent of time-aware applications that benefit from the knowledge of uncertainty in time, there is a need to rethink how time is managed in a system stack. Current clock models lack in their ability to expose different kind of uncertainties present in all the layers of a system stack, and time synchronization protocols do not provide any insight into their achieved performance, neither do they provide any hook to control their performance. In this paper, we present a new clock model exposing uncertainties in hardware, network stack and the operating system. Moreover, we implement a QoT-aware Kalman filter based synchronization protocol that makes uncertainty in time both observable and controllable.

We have left for future work to optimize network  and hardware resources with the desired timing accuracy, and test that we can conserve resources in terms of energy and bandwidth by making clock models and synchronization protocols QoT-aware.
\bibliographystyle{abbrv}
\bibliography{sigproc}  
%
%

\end{document}